\documentclass{article} 
\usepackage{iclr2026_conference,times}
\usepackage{booktabs}

\usepackage{amsmath,amsfonts,bm}

\def\eqref#1{equation~\ref{#1}}

\def\1{\bm{1}}

\DeclareMathAlphabet{\mathsfit}{\encodingdefault}{\sfdefault}{m}{sl}
\SetMathAlphabet{\mathsfit}{bold}{\encodingdefault}{\sfdefault}{bx}{n}

\usepackage{hyperref}
\usepackage{url}
\usepackage{graphicx}
\usepackage{wrapfig}
\usepackage{float}

\usepackage{algorithm}
\usepackage{algorithmic}

\title{FlowCast: Advancing Precipitation \\ Nowcasting with Conditional Flow Matching}

\author{Bernardo Perrone Ribeiro \\
University of Ljubljana\\
\texttt{brbmp0@gmail.com} \\
\And
Jana Faganeli Pucer \\
University of Ljubljana \\
\texttt{jana.faganelipucer@fri.uni-lj.si} \\
}

\iclrfinalcopy 
\begin{document}

\maketitle

\begin{abstract}
Radar-based precipitation nowcasting, the task of forecasting short-term precipitation fields from previous radar images, is a critical problem for flood risk management and decision-making. 
While deep learning has substantially advanced this field, two challenges remain fundamental: the uncertainty of atmospheric dynamics and the efficient modeling of high-dimensional data. 
Diffusion models have shown strong promise by producing sharp, reliable forecasts, but their iterative sampling process is computationally prohibitive for time-critical applications. 
We introduce FlowCast, the first end-to-end probabilistic model leveraging Conditional Flow Matching (CFM) as a direct noise-to-data generative framework for precipitation nowcasting. Unlike hybrid approaches, FlowCast learns a direct 
noise-to-data mapping in a compressed latent space, enabling rapid, high-fidelity sample generation. Our experiments demonstrate that FlowCast establishes a new state-of-the-art in 
probabilistic performance while also exceeding deterministic baselines in predictive accuracy. A direct comparison further reveals the CFM objective is both more accurate and significantly 
more efficient than a
diffusion objective on the same architecture, maintaining high performance with significantly fewer sampling steps. This work positions CFM as a powerful and practical alternative for 
high-dimensional spatiotemporal forecasting. \footnote{Our implementation is available at \url{https://github.com/b-rbmp/FlowCast}.}
\end{abstract}

\section{Introduction}
\label{sec:introduction}

Accurate and timely short-term precipitation forecasts, or nowcasting, are of paramount importance due to their significant socio-economic impacts, such as issuing flood warnings and managing water resources. 
Precipitation nowcasting, as defined in this work, involves predicting a sequence of future radar images from historical observations for the immediate future up to a few hours~\citep{an2024deep}. 
Traditional methods, like Eulerian and Lagrangian persistence~\citep{germann2002scale}, rely on advecting the current precipitation field. However, their simplified physical assumptions limit their ability to 
capture the complex, non-linear dynamics of atmospheric processes, especially for rapidly evolving weather systems~\citep{prudden2020review}.

Deep learning has introduced a paradigm shift in precipitation nowcasting. Deterministic models based on recurrent and transformer architectures learn complex spatiotemporal patterns directly from large volumes of 
radar data~\citep{prudden2020review, an2024deep}. While these models outperform traditional methods, optimizing for metrics like Mean Squared Error (MSE) compels them to produce a single, best-guess forecast. 
This often results in overly smooth predictions at longer lead times, failing to capture the inherent uncertainty in precipitation evolution and underrepresenting high-impact weather events.

To address this, probabilistic generative models have become central to modern nowcasting, aiming to predict a distribution over many plausible futures. Diffusion models~\citep{ho2020denoising}, in particular, 
have emerged as the state-of-the-art, producing sharp and reliable ensemble forecasts~\citep{gao2024prediff, leinonen2023latent, gong2024cascast}. However, their reliance on an 
iterative denoising process, often requiring hundreds of function evaluations for a single forecast, makes them computationally expensive. This high Number of Function Evaluations (NFE) poses a significant barrier in time-critical scenarios where rapid ensemble generation is crucial.
 
This work introduces FlowCast, a novel probabilistic nowcasting model built on Conditional Flow Matching (CFM)~\citep{lipman2023flow, tong2024improving}, a powerful and efficient alternative designed 
for rapid sampling. While recent work has applied rectified flows for the deterministic refinement of blurry forecasts~\citep{fengperceptually}, FlowCast is, to our knowledge, the first to 
successfully apply CFM as a full, noise-to-data generative model for this task. We demonstrate that FlowCast alleviates the tension between accuracy and efficiency, establishing a new 
state-of-the-art by exceeding the performance of leading diffusion models while offering a superior performance-cost trade-off.

We argue that CFM offers not only a computational advantage but also a superior inductive bias for this domain, specifically regarding the simplified transport of probability mass. 
Radar reflectivity distributions are highly multi-modal yet exhibit strong local temporal consistency. Standard diffusion models map Gaussian noise to this complex manifold via stochastic 
denoising or curved probability flow ODEs, often necessitating many sampling steps to resolve fine-grained structures without blurring modes. In contrast, CFM imposes a 
straight-line ODE prior on the generative process. This enforces the simplest possible mapping between the noise and data distributions. In the context of spatiotemporal forecasting, 
where temporal coherence is essential, this linear interpolation provides a much stronger and more stable prior than the winding paths of diffusion. We demonstrate that this geometric 
simplification allows FlowCast to maintain high fidelity with significantly fewer function evaluations.

Our contributions are summarized as follows:

\begin{itemize}
    \item We introduce FlowCast, a novel full-probabilistic application of Conditional Flow Matching to precipitation nowcasting.
    \item We establish a new state-of-the-art in both probabilistic performance and predictive accuracy on two diverse radar datasets, the benchmark SEVIR dataset~\citep{veillette2020sevir} and the local ARSO dataset.
    \item We provide a direct ablation study showing that the CFM objective is both more accurate and more computationally efficient than a diffusion objective on the same architecture, maintaining 
    high performance with substantially fewer sampling steps.
\end{itemize}

\section{Related Work}
\label{sec:related_work}
\subsection{Deterministic Nowcasting}
Deep learning for precipitation nowcasting has evolved from RNN-based architectures to Transformer-based models. Early work includes ConvLSTM~\citep{shi2015convolutional}, 
extending LSTMs with convolutions for spatiotemporal data, and the PredRNN family~\citep{wang2017predrnn, wang2022predrnn}, which introduced a spatiotemporal 
memory flow for improved long-range dependency modeling. More recently, Transformer architectures like Earthformer~\citep{gao2022earthformer} and 
Earthfarseer~\citep{wu2024earthfarsser} have set new benchmarks by using attention to model complex global dynamics. A common limitation of
deterministic models is that they produce overly smooth forecasts when trained with pixel-wise losses (e.g., MSE), as they average over possible futures.

\subsection{Probabilistic Nowcasting}
To address uncertainty quantification, probabilistic models have become central to nowcasting, aiming to sample from the full distribution of future states.

\paragraph{GANs and Diffusion.} 
GANs~\citep{ravuri2021skilful} were an early approach for producing sharp forecasts but suffer from training instability. Diffusion models~\citep{ho2020denoising} have recently 
emerged as the state-of-the-art, offering stable training and high-quality samples. PreDiff~\citep{gao2024prediff} and LDCast~\citep{leinonen2023latent} are prominent latent diffusion models 
for ensemble forecasting. A notable hybrid is CasCast~\citep{gong2024cascast}, which uses a deterministic model for large-scale patterns and a conditional 
diffusion model to refine stochastic details.

\paragraph{Flow-Based Generative Models.}
Generative modeling with flows offers an attractive alternative to diffusion. Traditional Continuous Normalizing Flows (CNFs)~\citep{chen2018neural} model data via ODEs but 
require expensive numerical integration during training to compute likelihoods, making them computationally prohibitive for high-dimensional spatiotemporal data. Conditional Flow Matching (CFM)~\citep{lipman2023flow} and Rectified Flows~\citep{liu2023flow} overcome this by regressing a vector field against a conditional probability path, 
enabling simulation-free training. However, while standard Rectified Flows typically utilize a singular conditional path (effectively $\sigma \to 0$), our application of Independent CFM (I-CFM) 
incorporates a Gaussian probability path with $\sigma > 0$. This "thickens" the training trajectory, providing crucial regularization that stabilizes the learning of the vector field 
for high-dimensional data compared to the singular paths of rectified flows.

Crucially, while diffusion models rely on stochastic denoising paths that are often curved and require many sampling steps, 
CFM allows for learning straight-line ODE trajectories between noise and data~\citep{tong2024improving}. This geometric property enforces a direct mapping that preserves temporal coherence 
and allows for rapid sampling. While \citet{fengperceptually} recently used a rectified flow module to strictly refine deterministic forecasts, 
FlowCast applies CFM as a standalone probabilistic generative model. This allows it to learn the full noise-to-data distribution and capture multimodal uncertainty without relying on a 
deterministic base forecast.

\section{Method}
\label{sec:method}

Our approach to probabilistic nowcasting is based on Conditional Flow Matching (CFM) within a compressed latent space. 
This section details our methodology, covering the problem formulation, our latent CFM framework, the model architecture, and the training and sampling procedures.

\subsection{Task Formulation}
\label{sec:task_formulation}
Precipitation nowcasting is framed as a video prediction task. Given a sequence of $T_{\mathrm{in}}$ past radar observations, 
$\mathbf{X}_{\mathrm{past}} = \{x_1, x_2, \ldots, x_{T_{\mathrm{in}}}\}$, where each $x_t \in \mathbb{R}^{H \times W \times C}$ is a radar 
map, the objective is to generate a probabilistic forecast for the next $T_{\mathrm{out}}$ frames, $\mathbf{X}_{\mathrm{future}} = \{x_{T_{\mathrm{in}}+1}, 
\ldots, x_{T_{\mathrm{in}}+T_{\mathrm{out}}}\}$.

\subsection{Latent Conditional Flow Matching}
\label{sec:latent_cfm}
To reduce the high computational cost of generative modeling, we adopt a two-stage approach inspired by latent diffusion 
models~\citep{rombach2022high}. A Variational Autoencoder (VAE)~\citep{kingma2013auto} compresses high-dimensional radar frames into 
low-dimensional latents, which are used to train a generative model in the latent space.

Our generative model is built on the Conditional Flow Matching (CFM) framework~\citep{lipman2023flow}, which trains a continuous 
normalizing flow by learning a vector field $v_\theta$ that maps samples from a prior distribution (e.g., Gaussian) 
to the target data distribution. We use Independent CFM (I-CFM)~\citep{tong2024improving}, which defines a probability 
path $p_t(x_t | x_0, x_1)$ as a Gaussian distribution with mean $(1-t)x_0 + t x_1$ and a small constant standard deviation $\sigma$. This path interpolates between a noise 
sample $x_0 \sim \mathcal{N}(0, I)$ and a data sample $x_1$. The corresponding target vector 
field is their difference, $u_t = x_1 - x_0$. This formulation enables direct, simulation-free training of the model 
$v_\theta$ by regressing it against this target field.

\subsubsection{Frame-wise Autoencoder}
\label{sec:autoencoder}
To learn a compact latent space, we train a VAE on individual radar frames. The architecture, inspired by \citet{esser2021taming}, uses a hierarchical 
encoder $\mathcal{E}$ and decoder $\mathcal{D}$ with residual and self-attention blocks for high-fidelity reconstructions. The VAE is trained with a combination of a L1 
reconstruction loss, a KL-divergence regularizer, and a PatchGAN adversarial loss~\citep{isola2017image} to enhance perceptual quality. After training, the VAE's weights 
are frozen and it is used to encode inputs and decode latent predictions.

\subsubsection{FlowCast Architecture}
\label{sec:flowcast_architecture}
\begin{figure}[htbp]
   \begin{center}
   \includegraphics[width=0.82\textwidth]{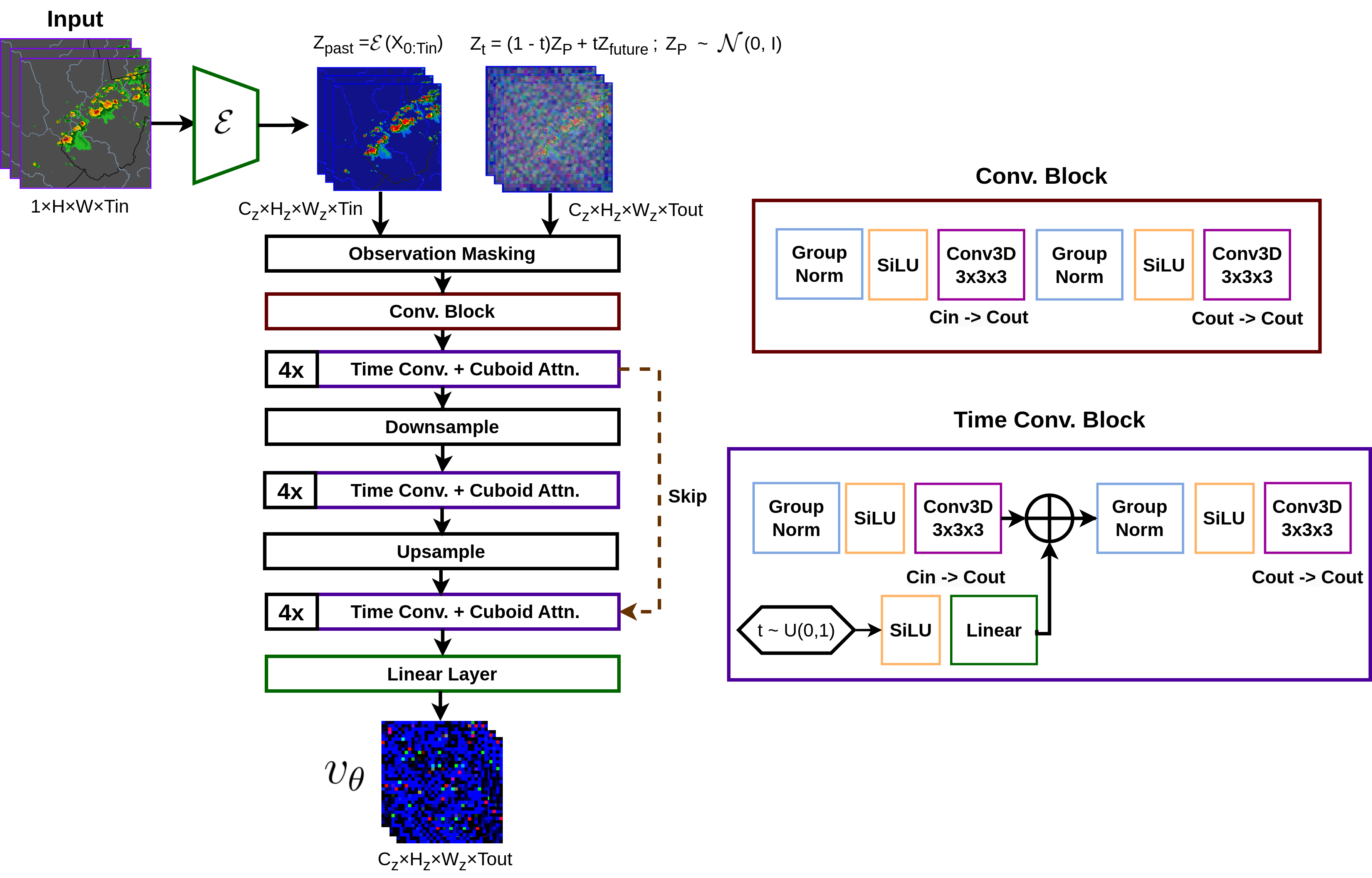}
   \end{center}
   \caption{The FlowCast architecture. A U-Net with Cuboid Attention blocks processes latent spatiotemporal data. Conditioning on the flow time $t$ enables the model to 
   learn the time-dependent vector field for generating forecasts.}
   \label{fig:flowcast}
\end{figure}

We propose \textbf{FlowCast}, which consists of the adaptation of Earthformer-UNet~\citep{gao2024prediff} for the CFM objective. FlowCast employs a 
U-Net-like encoder-decoder structure where the core building blocks are Cuboid Attention layers from 
Earthformer~\citep{gao2022earthformer}. This mechanism efficiently processes spatiotemporal data by applying self-attention locally 
within 3D "cuboids" of the data, capturing local dynamics, while global information is shared across the hierarchical U-Net structure. 
The model is conditioned on the flow time $t$, which is converted into an embedding and injected at each level of the network, 
enabling the model to accurately approximate the time-dependent vector field $v_\theta$. The architecture is illustrated in 
Figure~\ref{fig:flowcast}.

\paragraph{Training.} The FlowCast model learns the vector field $v_\theta(Z_t, t, Z_{\text{past}})$. The complete training procedure is detailed in Algorithm~\ref{alg:flowcast_training}.

\begin{algorithm}[htbp]
    \caption{FlowCast Training Process}
    \label{alg:flowcast_training}
    \begin{algorithmic}[1]
    \REQUIRE Dataset $\mathcal{D}$, Pre-trained VAE Encoder $\mathcal{E}$, FlowCast Model $v_\theta$, standard deviation $\sigma$.
    \STATE \textbf{Initialize} model parameters $\theta$
    \WHILE{not converged}
        \STATE Sample a batch of radar sequences $(X_{\text{past}}, X_{\text{future}}) \sim \mathcal{D}$
        \STATE Encode sequences into latent space: $Z_{\text{past}} \leftarrow \mathcal{E}(X_{\text{past}})$ and $Z_{\text{future}} \leftarrow \mathcal{E}(X_{\text{future}})$
        \STATE Sample prior noise $Z_P \sim \mathcal{N}(0, I)$, time $t \sim \mathcal{U}(0, 1)$ and path Gaussian noise $\epsilon \sim \mathcal{N}(0, I)$
        \STATE Compute interpolated latent state: $Z_t \leftarrow (1-t)Z_P + tZ_{\text{future}} + \sigma\epsilon$
        \STATE Compute target vector field: $u_t \leftarrow Z_{\text{future}} - Z_P$
        \STATE Predict vector field: $\hat{v} \leftarrow v_\theta(Z_t, t, Z_{\text{past}})$
        \STATE Compute Loss: $\mathcal{L} \leftarrow ||\hat{v} - u_t||^2$
        \STATE Gradient step: $\theta \leftarrow \theta - \eta \nabla_\theta \mathcal{L}$
    \ENDWHILE
    \end{algorithmic}
    \end{algorithm}

\paragraph{Sampling.} To generate an ensemble of forecasts, we solve the learned ODE starting from noise, using the Euler method~\citep{Hairer1993-za}. This process is described in Algorithm~\ref{alg:flowcast_sampling}.

\begin{algorithm}[htbp]
    \caption{FlowCast Ensemble Sampling}
    \label{alg:flowcast_sampling}
    \begin{algorithmic}[1]
    \REQUIRE Past radar sequence $X_{\text{past}}$, Trained FlowCast Model $v_\theta$, VAE Encoder $\mathcal{E}$ / Decoder $\mathcal{D}$.
    \REQUIRE Ensemble size $N$, Number of ODE steps $S$.
    \ENSURE Ensemble forecasts $\{\hat{X}_{\text{future}}^{(k)}\}_{k=1}^N$.
    \STATE Encode past observations: $Z_{\text{past}} \leftarrow \mathcal{E}(X_{\text{past}})$
    \FOR{$k = 1$ \textbf{to} $N$}
        \STATE Sample initial state: $Z(0) \sim \mathcal{N}(0, I)$
        \STATE Set step size: $\Delta t \leftarrow 1/S$
        \STATE Set current state: $Z_{curr} \leftarrow Z(0)$
        \FOR{$i = 0$ \textbf{to} $S-1$}
            \STATE Current time: $t \leftarrow i \cdot \Delta t$
            \STATE Predict vector field: $v \leftarrow v_\theta(Z_{curr}, t, Z_{\text{past}})$
            \STATE Update state (Euler step): $Z_{curr} \leftarrow Z_{curr} + v \cdot \Delta t$
        \ENDFOR
        \STATE Final latent forecast: $Z(1) \leftarrow Z_{curr}$
        \STATE Decode to pixel space: $\hat{X}_{\text{future}}^{(k)} \leftarrow \mathcal{D}(Z(1))$
    \ENDFOR
    \RETURN Ensemble predictions $\{\hat{X}_{\text{future}}^{(k)}\}_{k=1}^N$
\end{algorithmic}
\end{algorithm}

\section{Experiments}
\label{sec:experiments}

\subsection{Experimental Setting}
\label{sec:experimental_setting}
More details about the experimental setting are provided in Appendix~\ref{app:implementation_details}.

\subsubsection{Datasets}
\label{sec:datasets}

We evaluate on two 5-minute, 1 km-resolution radar datasets: SEVIR, a US benchmark, and ARSO, a Slovenian composite for a local deployment setting. For both, we 
predict 12 frames (1 hour) from 13 past frames (65 minutes), per the SEVIR Nowcasting Challenge protocol~\citep{veillette2020sevir}.

\begin{table}[htbp]
    \centering
    \caption{Summary of datasets used for evaluation.}
    \label{tab:datasetsummary}
    \bigskip
    \begin{tabular}{lccccccc}
    \toprule
    \textbf{Dataset} & \textbf{N\textsubscript{train}} & \textbf{N\textsubscript{val}} & \textbf{N\textsubscript{test}} & \textbf{Resolution} & \textbf{Dimensionality} & \textbf{Interval} & \textbf{Lag/Lead} \\
    \midrule
    SEVIR & 36,351 & 9,450 & 12,420 & 1 km & 384$\times$384 & 5 min & 13/12 \\
    ARSO & 38,229 & 12,743 & 12,744 & 1 km & 301$\times$401 & 5 min & 13/12 \\
    \bottomrule
    \end{tabular}
    
  \end{table}

\paragraph{SEVIR.}
\label{par:sevir_dataset}

SEVIR~\citep{veillette2020sevir} provides over 10,000 weather events in a 384$\times$384 km US domain, each spanning 4 hours at 5-minute resolution. We use the 1-km Vertically 
Integrated Liquid (VIL) field. Following the standard chronological split, we extract 25-frame sequences (13 context, 12 target) with a stride of 12, yielding 36,351 training, 
9,450 validation, and 12,420 test samples.

\paragraph{ARSO.}
\label{par:arso_dataset}

The ARSO dataset contains 5-minute, 1-km radar reflectivity composites over a 301$\times$401 km Slovenian grid, capturing complex Alpine and coastal dynamics. Using the 
same 25-frame sequence setup but with stride 1, a 60/20/20 chronological split yields 38,229 training, 12,743 validation, and 12,744 test samples.

\subsubsection{Evaluation}
\label{sec:evaluation}

\paragraph{Threshold-based categorical scores:}
Following prior work~\citep{veillette2020sevir,gao2024prediff,gong2024cascast}, we evaluate forecasts by converting radar fields to binary masks at given thresholds and computing 
the False Alarm Ratio (FAR), Critical Success Index (CSI), and Heidke Skill Score (HSS). For spatial validation, we compute the max-pooled CSI and Fractions Skill Score (FSS) over $16\times16$ km 
neighborhoods (CSI-M-P16 and FSS-M-P16). We report the mean of these scores across all thresholds ("-M") to evaluate general performance. Furthermore, to rigorously assess the detection of 
extreme weather events, we separately report the categorical metrics specifically at the highest intensity thresholds for each dataset.

For SEVIR, we follow the literature in using the thresholds $[16, 74, 133, 160, 181, 219]$. For ARSO, we use the thresholds $[15, 21, 30, 33, 36, 39]\,$dBZ, derived through quantile 
mapping to ensure that each threshold corres\-ponds to approximately the same exceedance probability in both datasets.

\paragraph{Continuous Ranked Probability Score (CRPS):}
We use the CRPS to evaluate probabilistic skill. A lower CRPS indicates a more accurate and sharp forecast. For deterministic forecasts ($N=1$), CRPS reduces to the 
Mean Absolute Error.

\paragraph{Ensemble forecasting:}
\label{p:ensemble_forecasting}
Let $x_{t,i,j}$ represent the ground truth pixel value at location $(i,j)$ and lead time $t$.
All probabilistic models are evaluated using an ensemble of $N=8$ realizations. For categorical scores, we evaluate the ensemble mean prediction
(Metric of Ensemble Mean), first computing the 
ensemble mean $\hat{x}_{t,i,j} = \frac{1}{N} \sum_{k=1}^{N} \hat{x}_{t,i,j}^{(k)}$ and then the metric on this mean forecast.

\subsubsection{Training Details}
\label{sec:training_details}

\paragraph{VAE.} We train a separate Variational Autoencoder (VAE) for each dataset to create a specialized latent space. We follow the architecture and training 
procedure from \citet{rombach2022high}, with a Kullback-Leibler divergence loss weight of 1e-4, the AdamW optimizer 
with a learning rate of 1e-4, and a batch size of 12. The compressed latent space dimensions are shown in Table~\ref{tab:latent_dims}.

\begin{table}[htbp]
    \centering
    \caption{VAE latent space dimensions}
    \label{tab:latent_dims}
    \bigskip
    \begin{tabular}{lcc}
        \toprule
        \textbf{Dataset} & \textbf{Original Dimensions} & \textbf{Latent Dimensions} \\
        & ($T_{\text{in}}/T_{\text{out}} \times \mathbf{H} \times \mathbf{W} \times \mathbf{C}_{\text{in}}$) & ($T_{\text{in}}/T_{\text{out}} \times \mathbf{H}_z \times \mathbf{W}_z \times \mathbf{C}_z$) \\
        \midrule
        SEVIR & $13/12 \times 384 \times 384 \times 1$ & $13/12 \times 48 \times 48 \times 4$ \\
        ARSO  & $13/12 \times 301 \times 401 \times 1$ & $13/12 \times 38 \times 52 \times 4$ \\
        \bottomrule
    \end{tabular}
\end{table}

\paragraph{FlowCast.} We train our CFM model for 200 epochs using the AdamW optimizer with a learning rate of 5e-4 and a cosine scheduler. We set the standard deviation of the I-CFM probability 
path to a small constant $\sigma=0.01$~\citep{tong2024improving}. We observed that the training process was notably stable; unlike diffusion objectives which often require complex loss weighting 
schedules, the I-CFM objective utilizes a simple regression loss that converged robustly without extensive hyperparameter tuning. Model checkpoints are maintained using an exponential moving 
average of weights~\citep{ho2020denoising}, with a decay factor of 0.999, and we keep the model checkpoint with
the highest CSI-M evaluated on a subset of the validation set. The model is trained with 4 NVIDIA H100 for 7 days, with a global batch size of 12. 
Further implementation details are 
provided in Appendix~\ref{app:implementation_details}.

\subsubsection{Inference Details}
\label{sec:inference_details}

Generating a forecast with FlowCast involves solving the learned ODE to transform a noise-initialized latent sequence into a prediction, conditioned on encoded past 
observations. Following the procedure outlined in~\ref{sec:flowcast_architecture}, we use the Euler method~\citep{Hairer1993-za} with 10 steps as the ODE solver. 
To generate a probabilistic ensemble forecast, this process is repeated eight times with different initial noise samples $Z(0) \sim \mathcal{N}(0, \mathbf{I})$.

\subsection{Comparison to the State of the Art}
\label{sec:sota}

We evaluate FlowCast against four deterministic baselines: U-Net~\citep{veillette2020sevir}, Earthformer~\citep{gao2022earthformer}, Earthfarseer~\citep{wu2024earthfarsser}, and SimVPv2~\citep{tan2025simvpv2}, 
as well as two probabilistic baselines: PreDiff~\citep{gao2024prediff} and CasCast~\citep{gong2024cascast}. All models are trained following their publicly released code, with 
the training budget fixed at 200 epochs. For probabilistic models, we adopt our evaluation protocol by selecting the checkpoint with the highest CSI-M on a validation subset, 
using exponential moving average weights.

\begin{figure}[H]
    \centering
    \includegraphics[width=0.95\textwidth]{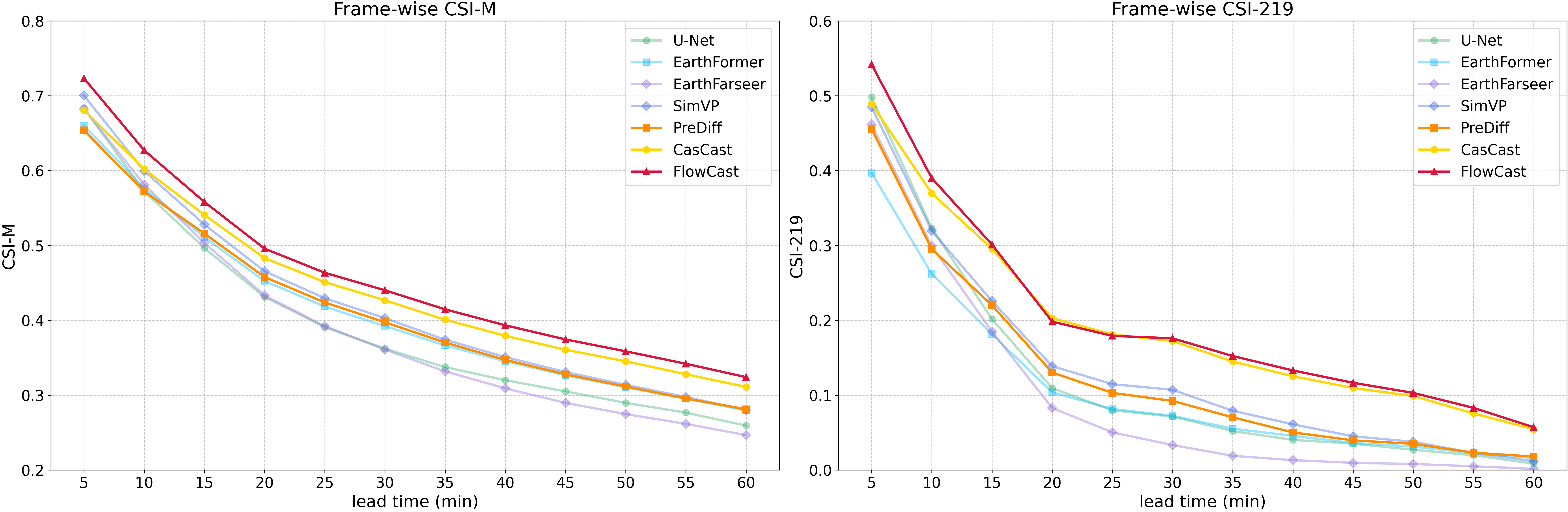}
    \caption{CSI-M and CSI at the 219 threshold per lead time on the SEVIR dataset. FlowCast shows consistent improvement over baselines for CSI-M and avoids the oversmoothing of 
    deterministic models at longer lead times (CSI-219).}
    \label{fig:sota_comparison_sevir_csi}
\end{figure}

\begin{figure}[htbp]
    \centering
    \includegraphics[width=0.99\textwidth]{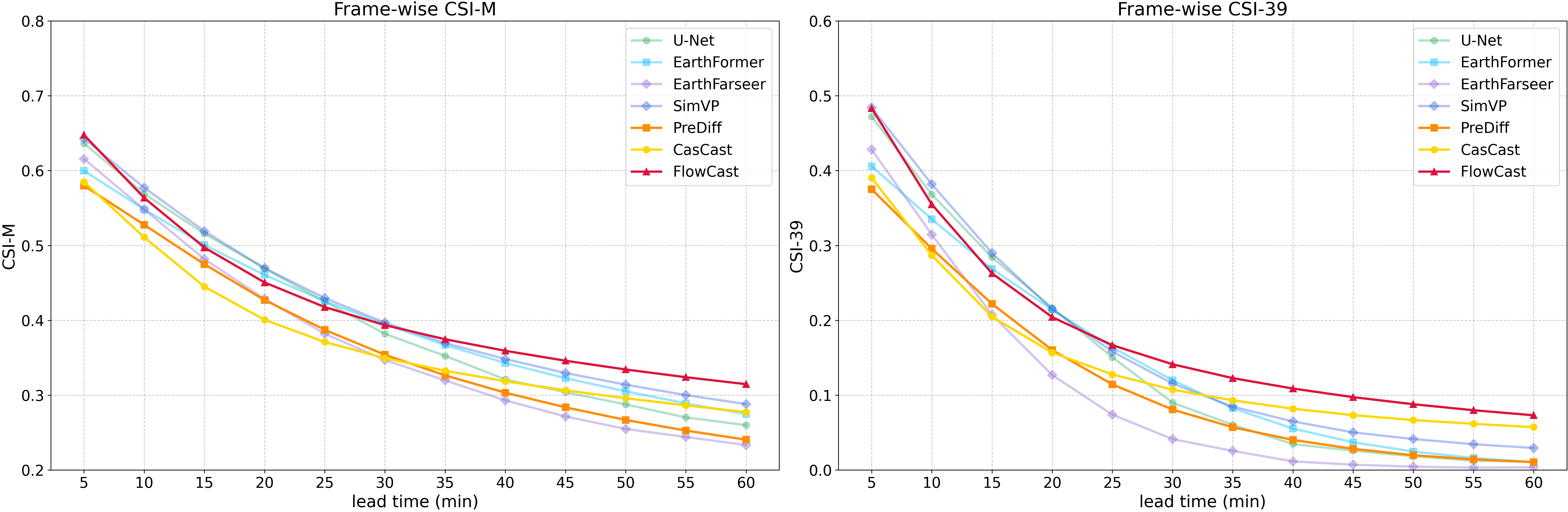}
    \caption{CSI-M and CSI at the 39 DBz threshold per lead time on the ARSO dataset. FlowCast shows significant improvements over probabilistic baselines for earlier lead times, and
    over deterministic baselines for later lead times.}
    \label{fig:sota_comparison_arso_csi}
\end{figure}

\textbf{General Performance (SEVIR \& ARSO).} As shown in Table~\ref{tab:sota_comparison_sevir}, Figure~\ref{fig:sota_comparison_sevir_csi}, Table~\ref{tab:sota_comparison_arso} and Figure~\ref{fig:sota_comparison_arso_csi}, FlowCast establishes a new state-of-the-art 
across both diverse datasets. On SEVIR, it achieves the highest overall CSI-M, FSS-M-P16, and HSS-M and the lowest CRPS, demonstrating superior probabilistic calibration. On ARSO, FlowCast 
outperforms all baselines in all metrics besides FAR-M. Notably, on SEVIR, while the probabilistic baseline CasCast achieves the highest CSI-P16-M, it suffers from a significantly 
higher FAR-M compared to FlowCast (0.383 vs. 0.325). This indicates that FlowCast strikes a superior balance between detection sensitivity and precision, 
avoiding the tendency to over-predict precipitation coverage. Deterministic models achieve the lowest FAR-M scores by predicting blurry fields at
longer lead times, missing the most extreme events (CSI-219).

\textbf{Performance on Extreme Events.} To evaluate the ability of the models to detect extreme events, we report performance at the highest intensity thresholds for both datasets in Table~\ref{tab:sota_comparison_sevir_arso_extreme_events}. 
FlowCast demonstrates a decisive advantage here. On SEVIR (Threshold 219), FlowCast achieves a CSI of 0.202, outperforming the best deterministic baseline (SimVP, 0.137) by over 47\% and 
the leading probabilistic baseline (CasCast, 0.195). The trend holds for ARSO (Threshold 39 dBZ), where FlowCast achieves the highest CSI (0.183) and HSS (0.291). Crucially, FlowCast maintains this 
high detection skill while achieving a lower FAR than CasCast across all extreme thresholds, confirming its ability to generate sharp, intense features without resorting to excessive false alarms.

\begin{table}[H]
    \centering
    \caption{Comparison of FlowCast with baseline models on the SEVIR dataset. All metrics are computed over a 12-step forecast, except "Forecast @ +65 min" which only uses the last frame.}
    
    \label{tab:sota_comparison_sevir}
    \bigskip
    \resizebox{\textwidth}{!}{
    \begin{tabular}{l|c|cc|c|c|c|cc}
        \toprule
        & & \multicolumn{5}{c|}{Forecast @ 12 steps} & \multicolumn{2}{c}{Forecast @ +65 min} \\
        \cmidrule(lr){3-7} \cmidrule(lr){8-9}
        Model & CRPS $\downarrow$ & CSI-M $\uparrow$ & CSI-P16-M $\uparrow$ & FSS-P16-M $\uparrow$ & HSS-M $\uparrow$ & FAR-M $\downarrow$ & CSI-M $\uparrow$ & CSI-219 $\uparrow$ \\
        \midrule
        U-Net & 0.0273 & 0.394 & 0.384 & 0.661 & 0.497 & 0.308 & 0.259 & 0.009 \\
        Earthformer & 0.0252 & 0.411 & 0.407 & 0.686 & 0.518 & \textbf{0.285} & 0.280 & 0.016 \\
        Earthfarseer & 0.0256 & 0.389 & 0.393 & 0.636 & 0.486 & 0.289 & 0.247 & 0.001 \\
        SimVP & 0.0249 & 0.423 & 0.424 & 0.701 & 0.532 & 0.298 & 0.280 & 0.012 \\
        \midrule
        PreDiff & 0.0189 & 0.413 & 0.423 & 0.699 & 0.523 & 0.313 & 0.281 & 0.018  \\
        CasCast & 0.0201 & 0.442 & \textbf{0.520} & 0.763 & 0.562 & 0.383 & 0.311 & 0.054 \\
        \midrule
        \textbf{FlowCast} & \textbf{0.0182} & \textbf{0.460} & 0.506 & \textbf{0.767} & \textbf{0.580} & 0.325 & \textbf{0.324} & \textbf{0.057} \\
        \bottomrule
    \end{tabular}
    }
\end{table}

\begin{table}[H]
    \centering
    \caption{Comparison of FlowCast with baseline models on the ARSO dataset. All metrics are computed over a 12-step forecast, except "Forecast @ +65 min" which only considers the last frame.}
    \label{tab:sota_comparison_arso}
    \bigskip
    \resizebox{\textwidth}{!}{
    \begin{tabular}{l|c|cc|c|c|c|cc}
        \toprule
        & & \multicolumn{5}{c|}{Forecast @ 12 steps} & \multicolumn{2}{c}{Forecast @ +65 min} \\
        \cmidrule(lr){3-7} \cmidrule(lr){8-9}
        Model & CRPS $\downarrow$ & CSI-M $\uparrow$ & CSI-P16-M $\uparrow$ & FSS-P16-M $\uparrow$ & HSS-M $\uparrow$ & FAR-M $\downarrow$ & CSI-M $\uparrow$ & CSI-39 $\uparrow$ \\
        \midrule
        U-Net & 0.0264 & 0.399 & 0.432 & 0.659 & 0.505 & 0.371 & 0.260 & 0.011 \\
        Earthformer & 0.0270 & 0.403 & 0.439 & 0.691 & 0.512 & 0.409 & 0.274 & 0.010 \\
        Earthfarseer & 0.0280 & 0.368 & 0.406 & 0.588 & 0.463 & \textbf{0.358} & 0.233 & 0.004 \\
        SimVP & 0.0267 & 0.415 & 0.462 & 0.699 & 0.526 & 0.401 & 0.288 & 0.029 \\
        \midrule
        PreDiff & 0.0211 & 0.369 & 0.411 & 0.614 & 0.471 & 0.400 & 0.241 & 0.010 \\
        CasCast & 0.0253 & 0.373 & 0.511 & 0.712 & 0.483 & 0.488 & 0.277 & 0.057 \\
        \midrule
        \textbf{FlowCast} & \textbf{0.0209} & \textbf{0.420} & \textbf{0.514} & \textbf{0.738} & \textbf{0.535} & 0.422 & \textbf{0.315} & \textbf{0.073} \\

        \bottomrule
    \end{tabular}
    }
\end{table}

\begin{table}[H]
    \centering
    \caption{Comparison of FlowCast with baseline models on SEVIR and ARSO datasets for extreme events using categorical metrics with the highest thresholds per dataset. All metrics are computed over a 12-step forecast.}
    \label{tab:sota_comparison_sevir_arso_extreme_events}
    \bigskip
    \resizebox{\textwidth}{!}{%
    \begin{tabular}{l | cc cc cc | cc cc cc}
        \toprule
        & \multicolumn{6}{c|}{SEVIR} & \multicolumn{6}{c}{ARSO} \\
        \cmidrule(lr){2-7} \cmidrule(lr){8-13}
        & \multicolumn{2}{c}{CSI} & \multicolumn{2}{c}{HSS} & \multicolumn{2}{c|}{FAR} & \multicolumn{2}{c}{CSI} & \multicolumn{2}{c}{HSS} & \multicolumn{2}{c}{FAR} \\
        \cmidrule(lr){2-3} \cmidrule(lr){4-5} \cmidrule(lr){6-7} \cmidrule(lr){8-9} \cmidrule(lr){10-11} \cmidrule(lr){12-13}
        Model & 181 & 219 & 181 & 219 & 181 & 219 & 36 & 39 & 36 & 39 & 36 & 39 \\
        \midrule
        U-Net & 0.205 & 0.122 & 0.314 & 0.193 & 0.366 & 0.508 & 0.209 & 0.145 & 0.318 & 0.226 & 0.474 & 0.505 \\
        Earthformer & 0.229 & 0.109 & 0.348 & 0.180 & 0.354 & \textbf{0.343} & 0.216 & 0.145 & 0.335 & 0.231 & 0.531 & 0.553 \\
        Earthfarseer & 0.194 & 0.097 & 0.291 & 0.152 & \textbf{0.341} & 0.412 & 0.159 & 0.104 & 0.245 & 0.164 & \textbf{0.445} & \textbf{0.449} \\
        SimVP & 0.244 & 0.137 & 0.365 & 0.220 & 0.370 & 0.404 & 0.238 & 0.162 & 0.362 & 0.254 & 0.507 & 0.540 \\
        \midrule
        PreDiff & 0.237 & 0.128 & 0.361 & 0.206 & 0.384 & 0.467 & 0.176 & 0.118 & 0.277 & 0.193 & 0.520 & 0.566 \\
        CasCast & 0.286 & 0.195 & 0.427 & 0.309 & 0.501 & 0.567 & 0.202 & 0.142 & 0.320 & 0.235 & 0.647 & 0.694 \\
        \midrule
        \textbf{FlowCast} & \textbf{0.301} & \textbf{0.202} & \textbf{0.443} & \textbf{0.317} & 0.425 & 0.482 & \textbf{0.254} & \textbf{0.183} & \textbf{0.388} & \textbf{0.291} & 0.547 & 0.589 \\
        \bottomrule
    \end{tabular}
    }
\end{table}

\begin{figure}[htbp]
    \centering
    \includegraphics[width=0.87\textwidth]{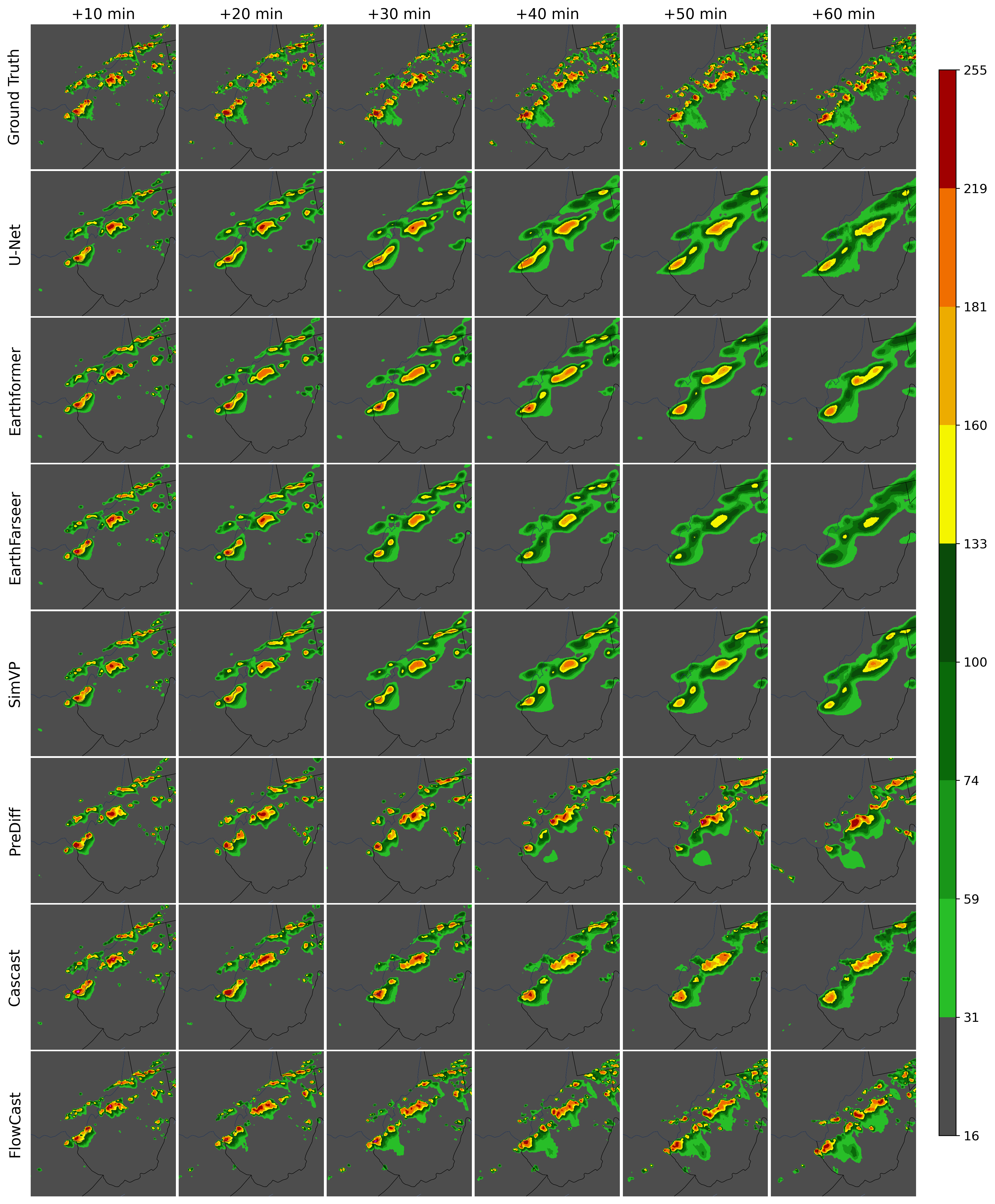}
    \caption{Qualitative comparison of FlowCast with other baselines on a SEVIR sequence. Columns show lead times from 10 to 60 minutes. Rows show the ground-truth, followed by the models.}
    \label{fig:sota_comparison_sevir}
\end{figure}

Figure~\ref{fig:sota_comparison_sevir} qualitatively compares forecast sequences from FlowCast with the baselines on the SEVIR dataset. FlowCast produces sharp, perceptually realistic forecasts, 
avoiding the smoothness of deterministic models. Compared to the best-performing probabilistic baseline CasCast, we observe more realistic precipitation patterns, especially at longer lead times. 
More examples, including on ARSO, are provided in Appendix~\ref{app:appendix_qualitative_examples}.

\subsection{Ablation Studies}
\label{sec:ablation}

Due to computational constraints, all ablation studies were run on the first 10\% of the SEVIR test set using a single NVIDIA A100 GPU.

\subsubsection{CFM objective against Diffusion}
\label{sec:ablation_cfm_vs_diffusion}

To isolate the benefits of the CFM objective, we compare FlowCast against a strong baseline using the same backbone architecture but trained with a diffusion objective. We trained a DDPM~\citep{ho2020denoising} for 1000 timesteps. For efficient inference, we employed a DDIM sampler~\citep{song2021denoising} with a varying number of steps. This provides a strong and 
practical baseline to evaluate FlowCast against a highly optimized diffusion process on the same powerful architecture.

The results in Table~\ref{tab:ablation_cfm_vs_diffusion} clearly demonstrate the superiority of the CFM objective. With a single step, FlowCast (CFM) drastically outperforms the DDIM sampler, even a 
100-step DDIM baseline, across key metrics like CRPS and CSI-M, whilst being almost 100 times faster.
\begin{table}[htbp]
    \centering
    \caption{Ablation study: CFM vs. diffusion objective. Results highlight the superior performance and efficiency of the CFM framework. All metrics are computed over a 12-step forecast.}
    \label{tab:ablation_cfm_vs_diffusion}
    \bigskip
    \resizebox{0.99\textwidth}{!}{
    \begin{tabular}{l|c|cc|c|c|c|c}
        \toprule
        Model & CRPS $\downarrow$ & CSI-M $\uparrow$ & CSI-P16-M $\uparrow$ & FSS-M-P16 $\uparrow$ & HSS-M $\uparrow$ & FAR-M $\downarrow$ & Time/Seq. (s) \\
        \midrule
        CFM (1 steps) & 0.0207 & 0.454 & 0.504 & 0.763 & 0.571 & 0.337 & \textbf{2.6} \\
        CFM (10 steps) & \textbf{0.0168} & \textbf{0.455} & \textbf{0.514} & \textbf{0.764} & \textbf{0.572} & 0.338 & 24 \\
        DDIM (10 steps) & 0.0262 & 0.395 & 0.450 & 0.622 & 0.503 & 0.335 & 24 \\
        DDIM (50 steps) & 0.0212 & 0.398 & 0.451 & 0.635 & 0.504 & 0.321 & 120 \\
        DDIM (100 steps) & 0.0208 & 0.398 & 0.450 & 0.664 & 0.502 & \textbf{0.319} & 239 \\
        \bottomrule
    \end{tabular}
    }
\end{table}

\subsubsection{Inference Efficiency: Performance vs. Number of Function Evaluations}
\label{sec:ablation_efficiency}

\begin{figure}[htbp]
    \centering
    \includegraphics[width=0.95\textwidth]{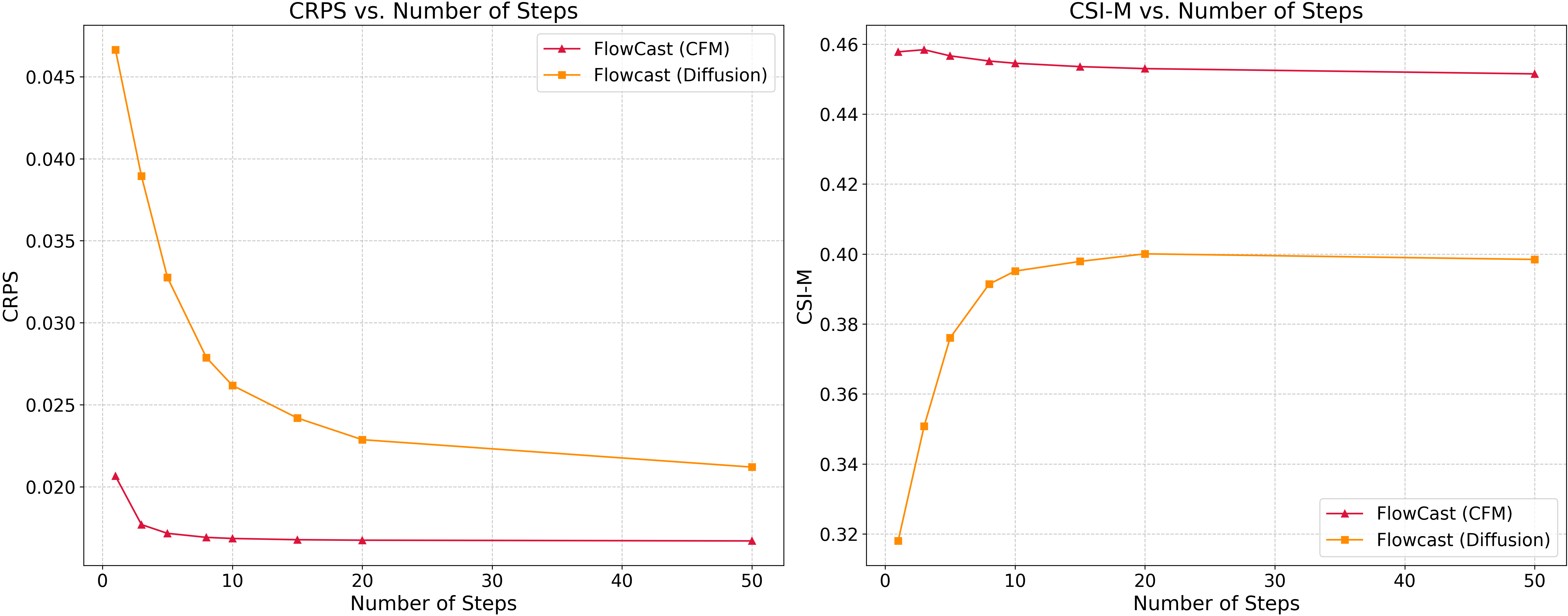}
    \caption{Performance vs. efficiency trade-off. Forecast quality (CRPS $\downarrow$, CSI-M $\uparrow$) as a function of NFE. FlowCast (CFM) 
    achieves near-optimal performance with only 3 to 10 steps, while the DDIM-based model requires 20 steps to 50 steps, and degrades sharply at low NFE.}
    \label{fig:ablation_nfe}
\end{figure}

We assess inference efficiency by comparing FlowCast and the diffusion backbone across a range of function evaluations (NFE), where one NFE is an Euler (CFM) or DDIM step. Each NFE adds ~2.4s per 8-member ensemble forecast. Figure~\ref{fig:ablation_nfe} shows FlowCast is highly efficient, 
nearing optimal CRPS and CSI-M scores in just 3-10 steps. In contrast, the diffusion model requires 20-50 steps to peak and degrades sharply below 10 NFE. These results highlight the superior efficiency of 
the CFM framework, which learns a more direct mapping to the data manifold and enables high-fidelity forecasts with significantly fewer model evaluations. This efficiency is a 
crucial advantage for operational settings where forecasts must be both rapid and reliable.

\section{Conclusion}
\label{sec:conclusion}

In this paper, we introduced FlowCast, the first fully probabilistic model applying Conditional Flow Matching (CFM) as a direct noise-to-data generative framework for precipitation nowcasting. Our experiments on the SEVIR and 
ARSO datasets show that FlowCast achieves state-of-the-art performance. Through direct ablation studies, we showed that the CFM objective is not only more accurate than a traditional diffusion objective on the same architecture 
but also vastly more efficient. FlowCast maintains high forecast quality with as few as a single sampling step, a regime where diffusion models fail. Our results firmly establish CFM as a powerful, efficient, and 
practical alternative for high-dimensional spatiotemporal forecasting.

\paragraph{Limitations and Future Work:}
\label{par:limitations}

While FlowCast shows significant promise, we identify two primary areas for future development. First, its reliance solely on radar data could be a limitation. Future work should explore 
multi-modal data fusion (e.g., satellite, NWP) to enhance robustness and accuracy. Second, our evaluation was 
limited to two datasets due to computational cost; a broader study across more meteorological regimes is needed to confirm generalizability.

\subsubsection*{Acknowledgments}
This work was supported by the Slovenian Research
and Innovation Agency (ARIS) research core funding P2-0209 (Jana Faganeli Pucer). Special thanks go to Janko Merše and Dr. Matic Šavli from the Slovenian Environment Agency (ARSO) for their support and for providing the ARSO dataset.

\subsubsection*{Reproducibility Statement}

To ensure the reproducibility of our work, we have provided comprehensive supporting materials. \textbf{Code:} The full source code for our FlowCast model, including scripts for 
training and evaluation, is available at \url{https://github.com/b-rbmp/FlowCast}. The repository contains detailed instructions for setting up the required software environment and running the 
experiments. \textbf{Datasets:} Our work utilizes two datasets. The SEVIR dataset is a public benchmark, and details for access are provided by \citet{veillette2020sevir}. The ARSO 
dataset was provided by the Slovenian Environment Agency (ARSO) for this research; we are actively collaborating with the agency to facilitate its public release in the near future. 
\textbf{Experimental Details:} Section~\ref{sec:experimental_setting} of the main paper provides a detailed description of our experimental setup. Further implementation details are available in Appendix~\ref{app:implementation_details}, including
all model hyperparameters.

\bibliography{iclr2026_conference}
\bibliographystyle{iclr2026_conference}

\clearpage
\appendix
\section{Appendix}
\label{sec:appendix}

\subsection{Implementation Details}
\label{app:implementation_details}

This section details the implementation of FlowCast. The code to train and evaluate FlowCast is provided as supplementary material.

\subsubsection{VAE}
\label{app:vae_training}

The VAE was configured with the hyperparameters shown in Table~\ref{tab:vae_training_summary}. Different warmup periods were empirically selected based on the convergence speed of the VAE on the 
respective datasets; ARSO converged faster due to less diversity in the data, attributed to factors such as its fixed geographical coverage and the significantly smaller stride used 
during sequence extraction.

\begin{table}[H]
    \centering
    \caption{VAE hyperparameter summary}
    \label{tab:vae_training_summary}
    \bigskip
    \resizebox{\textwidth}{!}{%
    \begin{tabular}{@{}lll@{}}
    \toprule
    \textbf{Category} & \textbf{Parameter} & \textbf{Value / Setting} \\
    \midrule
    \multicolumn{3}{l}{\textit{\textbf{Dataset}}} \\
    & Source & SEVIR (\texttt{vil}) \& ARSO (\texttt{zm}) \\
    & Input Dimensionality (per frame) & 384 $\times$ 384 $\times$ 1 (SEVIR), 301 $\times$ 401 $\times$ 1 (ARSO) \\
    & Input Preprocessing & Frame values scaled to $[0, 1]$ \\
    \midrule
    \multicolumn{3}{l}{\textit{\textbf{Training Objective}}} \\
    & Loss Components & Reconstruction + KL Divergence + Adversarial \\
    & KL Divergence Weight ($\lambda_{KL}$) & $1 \times 10^{-4}$ \\
    & Discriminator Weight ($\lambda_{adv}$) & 0.5 \\
    & Adversarial Loss Type & Hinge Loss \\
    & Discriminator Architecture & PatchGAN~\citep{isola2017image} \\
    & Discriminator Activation Warmup & 35 epochs (SEVIR), 15 epochs (ARSO) \\
    \midrule
    \multicolumn{3}{l}{\textit{\textbf{Optimization}}} \\
    & Optimizer (Generator \& Disc.) & AdamW \\
    & Learning Rate (Initial) & $1 \times 10^{-4}$ \\
    & Weight Decay & $1 \times 10^{-5}$ \\
    & AdamW Betas & $(0.9, 0.999)$ \\
    & LR Scheduler & Cosine Annealing with Linear Warmup \\
    & LR Warmup Fraction & 20\% of total training steps \\
    & LR Min Warmup Ratio & 0.1 \\
    & Min. LR Ratio & $10^{-3}$ \\
    \midrule
    \multicolumn{3}{l}{\textit{\textbf{Training Configuration}}} \\
    & Batch Size & 12 (Global), 3 (Local) \\
    & Max. Number of Epochs & 250 \\
    & Gradient Clipping Norm & 1.0 \\
    & Early Stopping Patience & 50 epochs \\
    & Early Stopping Metric & Generator validation loss \\
    & Training Nodes & 4 $\times$ H100 GPUs \\
    & FP16 Training & Disabled \\
    \midrule
    \multicolumn{3}{l}{\textit{\textbf{Model Configuration}}} \\
    & Latent Channels & 4 \\
    & GroupNorm Num & 32 \\
    & Layers per Block & 2 \\
    & Activation Function & SiLU \\
    & Encoder-Decoder Depth & 4 \\
    & Block Out Channels & [128, 256, 512, 512] \\
    \midrule
    \bottomrule
    \end{tabular}
    }
\end{table}

\paragraph{Data Preprocessing and Padding.} 
To accommodate the VAE's downsampling factor of $f=8$, inputs must be spatially divisible by the downsampling rate. For the ARSO dataset, the native resolution of $301 \times 401$ is not divisible 
by 8. We handle this by applying replication padding to the input frames to reach the nearest multiple of 16 prior to encoding. Specifically, the height is padded from 301 to 304, and the 
width from 401 to 416. This results in the latent dimensions of $38 \times 52$ reported in Table~\ref{tab:latent_dims} ($304/8$ and $416/8$, respectively). During inference, 
the generated fields are cropped back to the original $301 \times 401$ dimensions before evaluation metrics are computed. SEVIR dimensions ($384 \times 384$) are naturally divisible by 8, 
requiring no padding.

\subsubsection{FlowCast}
\label{app:flowcast_architecture}

\paragraph{Architecture.} The FlowCast architecture, adapted from Earthformer-UNet~\citep{gao2024prediff} for latent-space Conditional Flow Matching, has the following configuration:
\begin{itemize}
    \item \textbf{Core U-Net Architecture:}
        \begin{itemize}
            \item \textbf{Hierarchical Stages:} A U-Net with 2 hierarchical stages 
            (one level of downsampling/upsampling within the main U-Net body, in addition to initial/final processing).
            \item \textbf{Stacked Cuboid Self-Attention Modules:} Each stage in both the contracting (encoder) and expansive (decoder) paths contains a depth of 
            4 Stacked Cuboid Self-Attention modules.
            \item \textbf{Base Feature Dimensionality:} 192 units.
        \end{itemize}
    \item \textbf{Spatial Processing:}
        \begin{itemize}
            \item \textbf{Downsampling:} Achieved using Patch Merge (reducing spatial dimensions by a factor of 2 and doubling channel depth).
            \item \textbf{Upsampling:} Uses nearest-neighbor interpolation followed by a convolution (halving channel depth).
        \end{itemize}
    \item \textbf{Cuboid Self-Attention Details:}
        \begin{itemize}
            \item \textbf{Pattern:} Follows an axial pattern, processing temporal, height, and width dimensions sequentially.
            \item \textbf{Attention Heads:} 4 attention heads.
            \item \textbf{Positional Embeddings:} Relative positional embeddings are used.
            \item \textbf{Projection Layer:} A final projection layer is part of the attention block.
            \item \textbf{Dropout Rates:} Dropout rates for attention, projection, and Feed-Forward Network (FFN) layers are set to 0.1.
        \end{itemize}
    \item \textbf{Global Vectors:} The specialized global vector mechanism from the original Earthformer is disabled.
    \item \textbf{FFN and Normalization:}
        \begin{itemize}
            \item \textbf{FFN Activation:} Feed-forward networks within the attention blocks use GELU activation.
            \item \textbf{Normalization:} Layer normalization is applied throughout the relevant parts of the network.
        \end{itemize}
    \item \textbf{Embeddings:}
        \begin{itemize}
            \item \textbf{Spatiotemporal Positional Embeddings:} Added to the input features after an initial projection.
            \item \textbf{CFM Time Embeddings ($t$):}
                \begin{itemize}
                    \item \textbf{Generation:} Generated with a channel multiplier of 4 relative to the base feature dimensionality 
                    (resulting in $192 \times 4 = 768$ embedding channels).
                    \item \textbf{Incorporation:} Injected into the network at each U-Net stage using residual blocks that fuse the time embedding with the feature maps (\texttt{TimeEmbedResBlock} modules). 
                \end{itemize}
        \end{itemize}
    \item \textbf{Skip Connections:} Standard U-Net additive skip connections merge features from the contracting path to the expansive path.
    \item \textbf{Padding:} Zero-padding is used where necessary to maintain tensor dimensions during convolutions or cuboid operations.
\end{itemize}

\paragraph{Training and Inference Hyperparameters.} The FlowCast training and inference hyperparameters are as follows:
\begin{table}[H]
    \centering
    \caption{FlowCast hyperparameter summary}
    \label{tab:flowcast_training_summary}
    \bigskip
    \resizebox{\textwidth}{!}{%
    \begin{tabular}{@{}lll@{}}
    \toprule
    \textbf{Category} & \textbf{Parameter} & \textbf{Value / Setting} \\
    \midrule
    \multicolumn{3}{l}{\textit{\textbf{Dataset}}} \\
    & Source & Latent-space sequences from VAE \\
    & Input Dimensionality & 13 $\times$ 48 $\times$ 48 $\times$ 4 (SEVIR), 13 $\times$ 38 $\times$ 52 $\times$ 4 (ARSO) \\
    & Input Preprocessing & Standardized with training set statistics (mean, std) \\
    \midrule
    \multicolumn{3}{l}{\textit{\textbf{Training Configuration}}} \\
    & Loss Function & MSE: $\mathcal{L} = \| \hat{v} - (Z_{\text{future}} - Z_P) \|^2$ \\
    & Batch Size & 12 (Global), 3 (Local) \\
    & Max. Number of Epochs & 200 \\
    & Gradient Clipping Norm & 1.0 \\
    & Early Stopping Patience & 50 epochs \\
    & Early Stopping Metric & CSI-M evaluated on subset (40 batches) of validation set \\
    & Training Nodes & 4 $\times$ H100 GPUs \\
    & FP16 Training & Enabled \\
    & Exponential Moving Average Weights & Enabled \\
    & Exponential Moving Average Weights Decay & 0.999 \\
    \midrule
    \multicolumn{3}{l}{\textit{\textbf{Optimization}}} \\
    & Optimizer & AdamW \\
    & Learning Rate (Initial) & $5 \times 10^{-4}$ \\
    & Weight Decay & $1 \times 10^{-4}$ \\
    & AdamW Betas & $(0.9, 0.999)$ \\
    & LR Scheduler & Cosine Annealing with Linear Warmup \\
    & LR Warmup Fraction & 1\% of total training steps \\
    & LR Min Warmup Ratio & 0.1 \\
    & Min. LR Ratio & $10^{-2}$ \\
    \midrule
    \multicolumn{3}{l}{\textit{\textbf{CFM Parameters}}} \\
    & $\sigma$ & 0.01 \\
    & ODE Solver & Euler Method with 10 steps \\
    \bottomrule
    \end{tabular}
    }
\end{table}

\paragraph{Choice of ODE Solver.} We conducted an ablation study to compare various ODE solvers, including adaptive methods (Adaptive Heun, Dormand-Prince 5) and fixed-step methods 
(Euler, Midpoint, Runge-Kutta 4) on the first 10\% of the SEVIR test set using a single NVIDIA A100 GPU. For adaptive solvers, a relative and absolute tolerance of $10^{-2}$ and $10^{-3}$ were used, respectively. Since no 
significant performance differences were observed, as shown in Table~\ref{tab:ablation_ode_solvers}, we selected the Euler method with 10 steps for its computational efficiency and simplicity.

\begin{table}[htbp]
    \centering
    \caption{Ablation study: ODE solvers. Results highlight minor differences in performance between the different solvers.}
    \label{tab:ablation_ode_solvers}
    \bigskip
    \resizebox{0.99\textwidth}{!}{
    \begin{tabular}{l|c|cc|c|c|c|c}
        \toprule
        Solver & CRPS $\downarrow$ & CSI-M $\uparrow$ & CSI-P16-M $\uparrow$ & FSS-M-P16 $\uparrow$ & HSS-M $\uparrow$ & FAR-M $\downarrow$ & Time/Seq. (s) \\
        \midrule
        Euler (1 steps) & 0.0207 & 0.454 & 0.504 & 0.763 & 0.571 & 0.337 & 2.6 \\
        Euler (10 steps) & 0.0168 & 0.455 & 0.514 & 0.764 & 0.572 & 0.338 & 24 \\
        Dormand-Prince 5 & 0.0168 & 0.450 & 0.516 & 0.762 & 0.567 & 0.341 & 46 \\
        Midpoint (10 steps) & 0.0167 & 0.451 & 0.516 & 0.762 & 0.568 & 0.341 & 44 \\
        Runge-Kutta 4 (10 steps) & 0.0167 & 0.451 & 0.516 & 0.762 & 0.567 & 0.341 & 83  \\
        Adaptive Heun & 0.0167 & 0.451 & 0.516 & 0.762 & 0.568 & 0.341 & 50 \\
        \bottomrule
    \end{tabular}
    }
\end{table}

\subsubsection{Evaluation Metrics}
\label{app:evaluation_metrics}

\paragraph{Continuous Ranked Probability Score (CRPS).}

The CRPS is evaluated directly at the original data resolution, without applying any spatial pooling.
For each ensemble of $N$ forecast members, CRPS is calculated at every pixel and then averaged across all spatial positions and forecast lead times to obtain a single summary metric.
If the predictive distribution $F$ at a given pixel and time step is approximated by a Gaussian with mean $\mu$ and standard deviation $\sigma$ (estimated from the ensemble), and $x$ is the observed value, the CRPS can be computed as:

\begin{equation}
    \text{CRPS}(F, x) = \sigma \left( \frac{x-\mu}{\sigma} \left(2\Phi\left(\frac{x-\mu}{\sigma}\right) - 1\right) + 2\phi\left(\frac{x-\mu}{\sigma}\right) - \frac{1}{\sqrt{\pi}} \right),
\end{equation}

where $\Phi$ and $\phi$ denote the cumulative distribution function (CDF) and probability density function (PDF) of the standard normal distribution, respectively. Following standard practice, we normalize the CRPS for each dataset by its maximum values, which scales our reported metrics by 1/255 for SEVIR and 1/57 for ARSO.

\paragraph{Threshold-based categorical metrics.}

For each chosen intensity threshold $u$, we binarize the continuous ground truth field $x_{t,i,j}$ to obtain an observation mask $\mathbb{1}[x_{t,i,j} > u]$. The representative 
forecast $\hat{x}_{t,i,j}$ (see Section~\ref{p:ensemble_forecasting}) is likewise thresholded to produce a binary forecast mask $\mathbb{1}[\hat{x}_{t,i,j} > u]$.

Using these binary masks and the dataset-specific thresholds, we construct a $2\times2$ contingency table for each evaluation:
\[
\begin{array}{c|cc}
         & \text{Observ. yes} & \text{Observ. no}\\ \hline
\text{Forecast yes} & H & F \\
\text{Forecast no } & M & C
\end{array}
\qquad
\begin{aligned}
H &: \text{Hits (True Positives)},\\
M &: \text{Misses (False Negatives)},\\
F &: \text{False Alarms (False Positives)},\\
C &: \text{Correct Negatives}.
\end{aligned}
\]
The values $H, M, F, C$ are summed over all spatial locations, batches, and forecast lead times. The following metrics are then computed:

\paragraph{False Alarm Ratio (FAR):} The proportion of forecasted events that did not actually occur.

\begin{equation}
    \mathrm{FAR} = \frac{F}{H + F}.
\end{equation}

\paragraph{Critical Success Index (CSI):} The fraction of observed and/or forecasted events that were correctly predicted, ignoring correct negatives. This metric is sensitive to both missed events and false alarms.
\begin{equation}
    \mathrm{CSI} = \frac{H}{H + M + F}.
\end{equation}

\paragraph{Heidke Skill Score (HSS):} The accuracy of the forecast relative to random chance.
\begin{equation}
    \mathrm{HSS} = 
    \frac{2(HC - MF)}{
    (H + M)(M + C) + (H + F)(F + C)
    }.
\end{equation}

\paragraph{Pooled CSI (CSI-P16):}
Forecasts and ground truth fields are first downsampled by applying a max-pooling operation over non-overlapping $16 \times 16$ pixel blocks. CSI is then recomputed on these 
pooled fields. A hit anywhere within a $16 \times 16$ block is registered as a success, thereby rewarding models that capture the local presence and intensity of 
precipitation, even if exact pixel-level alignment is imperfect.

\paragraph{Fractions Skill Score (FSS):}
We compute FSS to evaluate spatial alignment. Let $S_f(i,j)$ and $S_o(i,j)$ denote the fraction of pixels exceeding the threshold within a neighborhood of size $n \times n$ centered at $(i,j)$ for the forecast and observation, respectively. The FSS is calculated as:
\begin{equation}
    \mathrm{FSS} = 1 - \frac{\sum_{i,j} [S_f(i,j) - S_o(i,j)]^2}{\sum_{i,j} S_f(i,j)^2 + \sum_{i,j} S_o(i,j)^2}
\end{equation}

\subsection{More Qualitative Examples}
\label{app:appendix_qualitative_examples}

This section presents additional qualitative examples comparing FlowCast against the baseline models

\subsubsection{SEVIR}
\label{app:sevir_examples}

\begin{figure}[H]
    \centering
    \vspace*{\fill}
    \includegraphics[width=0.99\textwidth]{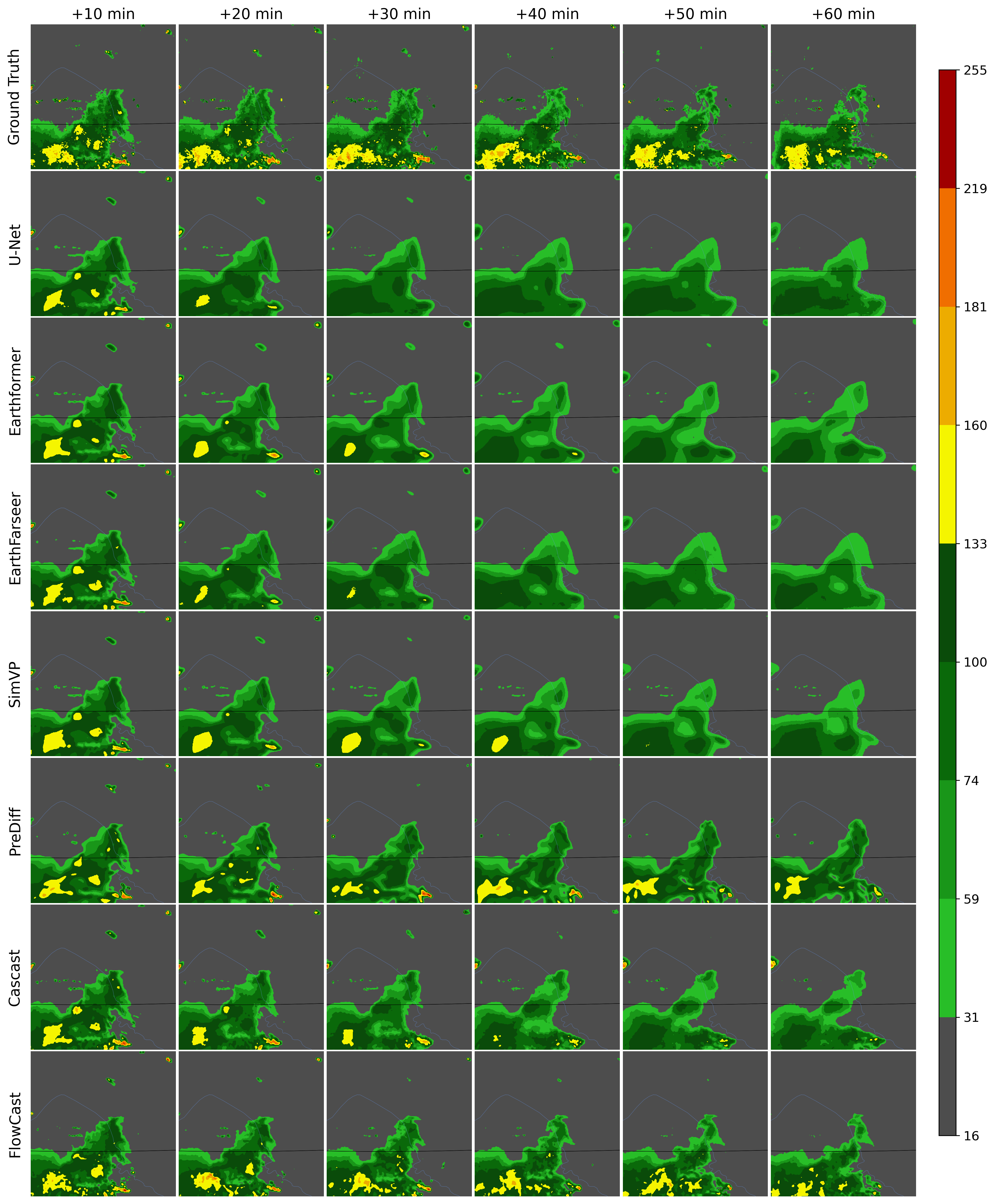}
    \caption{Qualitative comparison of FlowCast with other baselines on a SEVIR sequence.}
    \label{fig:sevir_examples_1}
    \vspace*{\fill}
\end{figure}
\clearpage

\begin{figure}[p]
    \centering
    \vspace*{\fill}
    \includegraphics[width=0.99\textwidth]{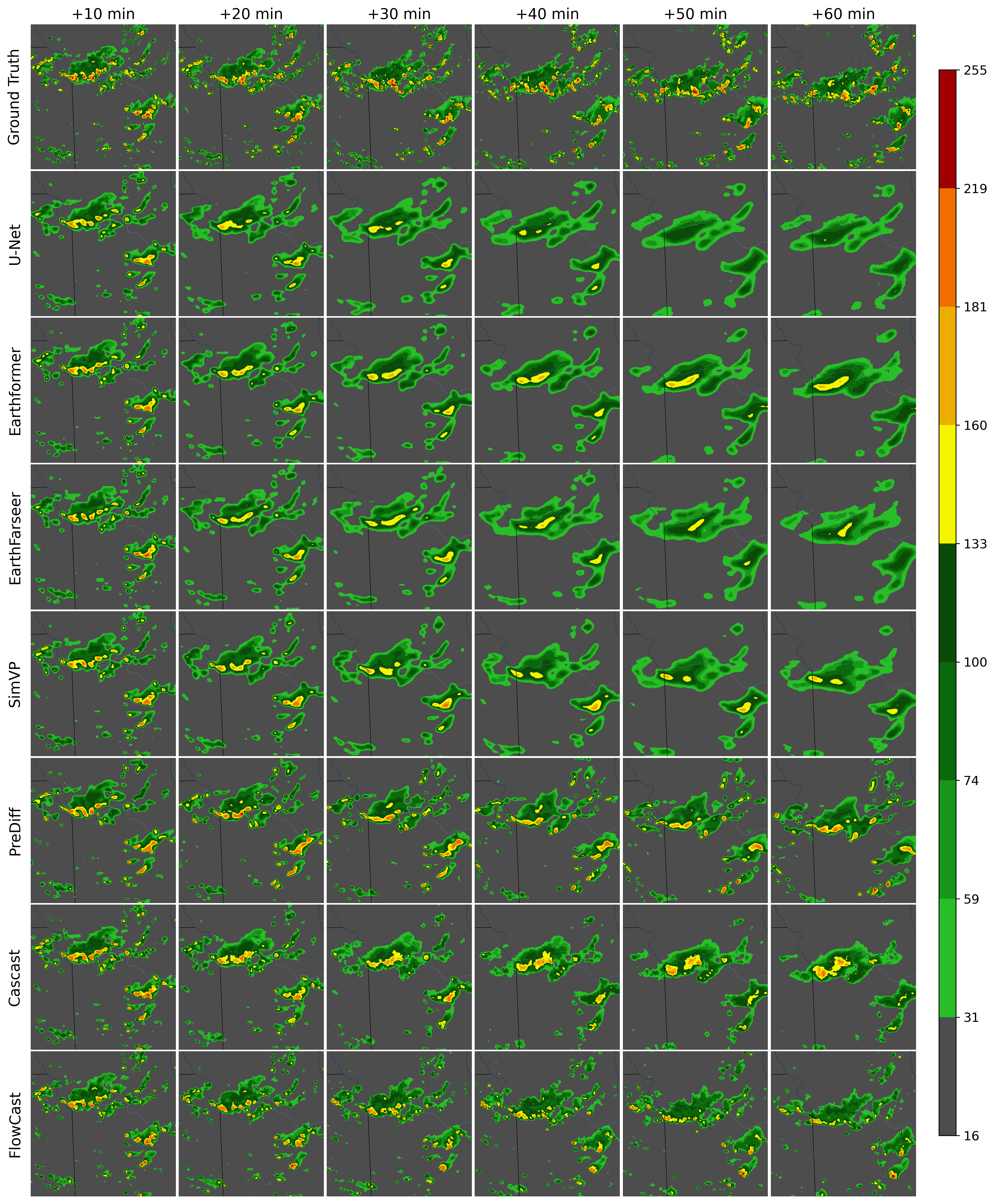}
    \caption{Qualitative comparison of FlowCast with other baselines on a SEVIR sequence.}
    \label{fig:sevir_examples_2}
    \vspace*{\fill}
\end{figure}
\clearpage

\begin{figure}[p]
    \centering
    \vspace*{\fill}
    \includegraphics[width=0.99\textwidth]{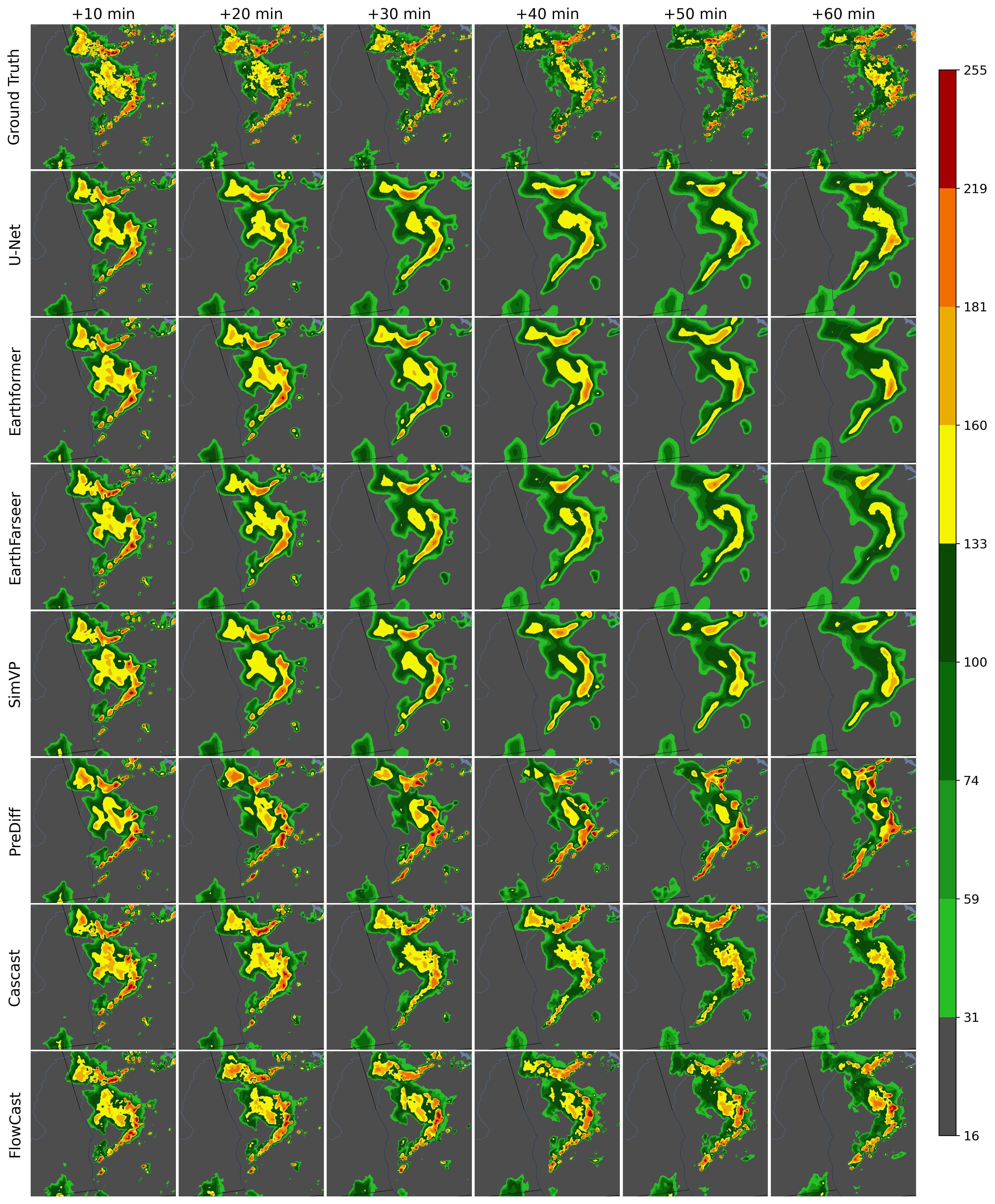}
    \caption{Qualitative comparison of FlowCast with other baselines on a SEVIR sequence.}
    \label{fig:sevir_examples_3}
    \vspace*{\fill}
\end{figure}
\clearpage

\subsubsection{ARSO}
\label{app:arso_examples}

\begin{figure}[H]
    \centering
    \vspace*{\fill}
    \includegraphics[width=0.99\textwidth]{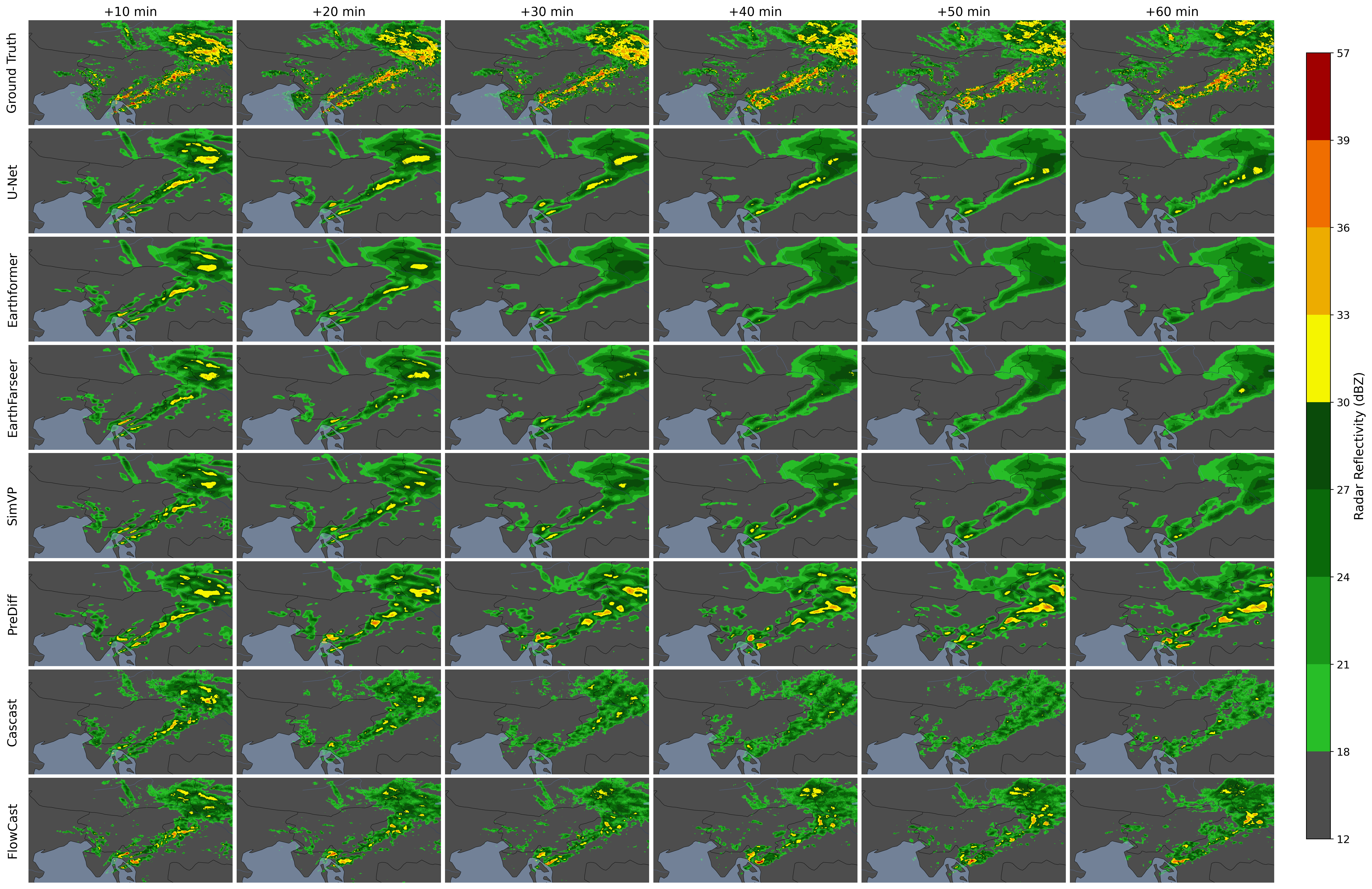}
    \caption{Qualitative comparison of FlowCast with other baselines on an ARSO sequence.}
    \label{fig:arso_examples_1}
    \vspace*{\fill}
\end{figure}

\begin{figure}[H]
    \centering
    \vspace*{\fill}
    \includegraphics[width=0.99\textwidth]{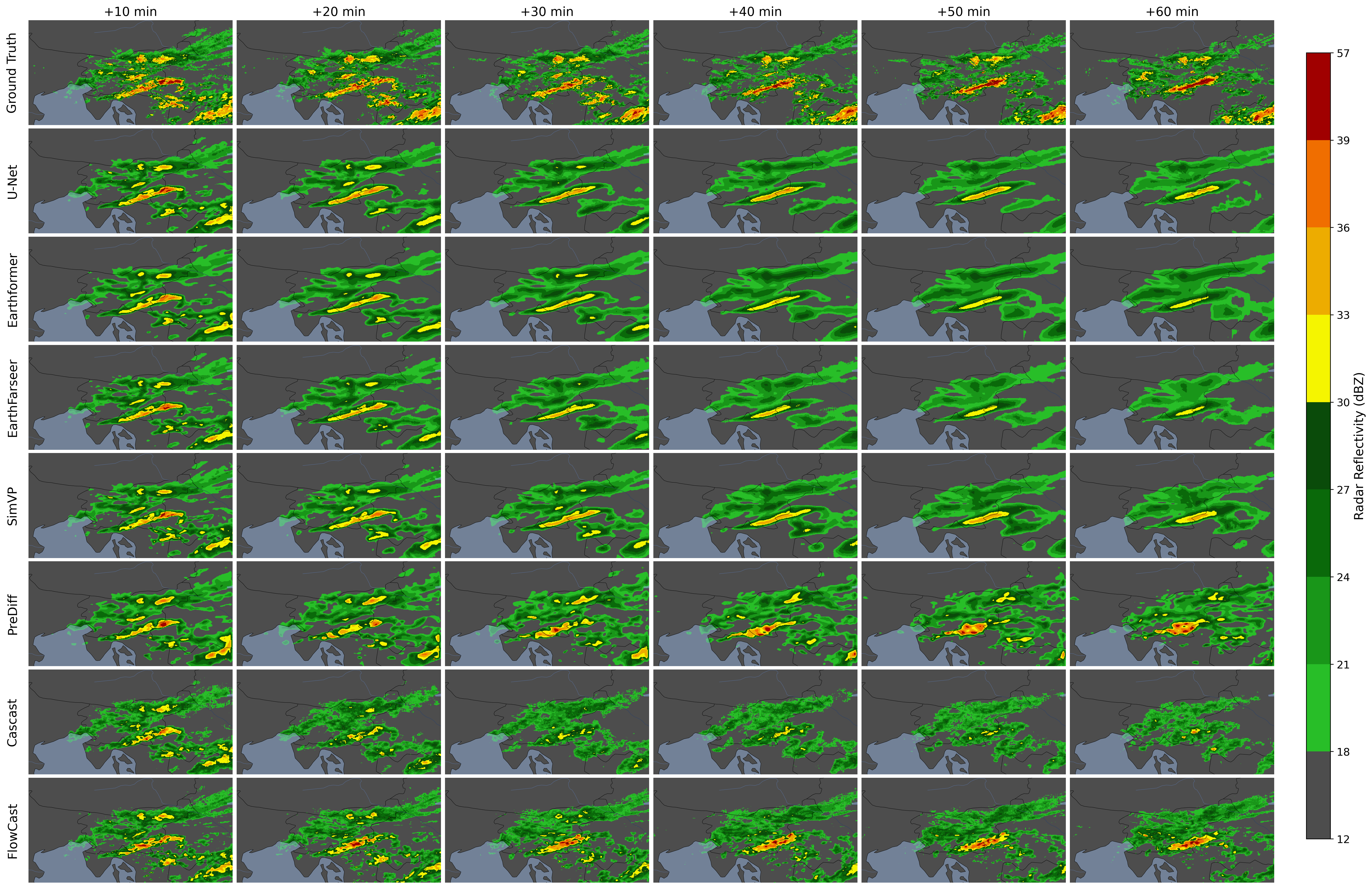}
    \caption{Qualitative comparison of FlowCast with other baselines on an ARSO sequence.}
    \label{fig:arso_examples_2}
    \vspace*{\fill}
\end{figure}

\begin{figure}[H]
    \centering
    \vspace*{\fill}
    \includegraphics[width=0.99\textwidth]{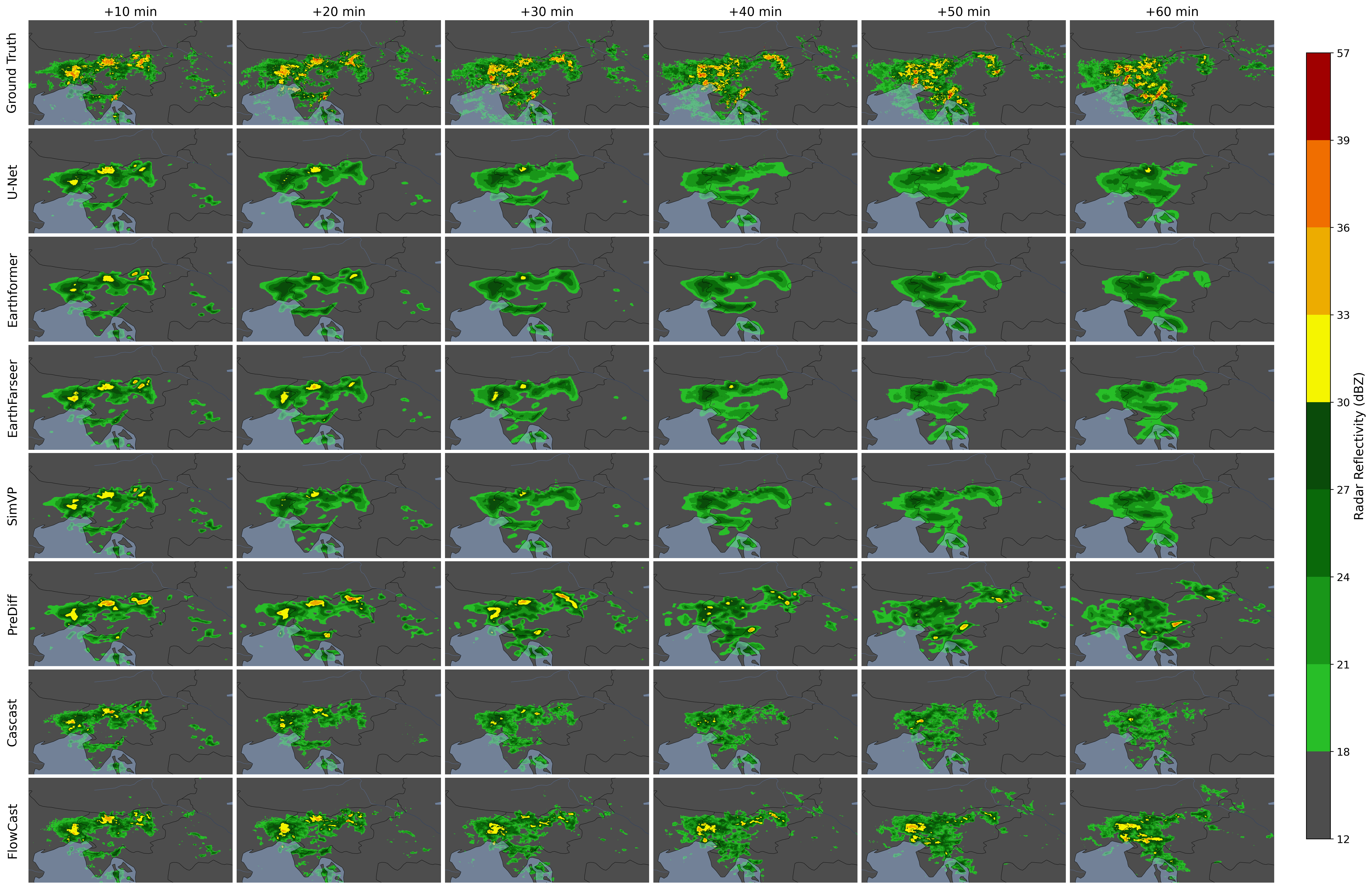}
    \caption{Qualitative comparison of FlowCast with other baselines on an ARSO sequence.}
    \label{fig:arso_examples_3}
    \vspace*{\fill}
\end{figure}

\subsection{LLM Usage Statement}

During the preparation of this manuscript, we utilized a large language model (LLM) as a writing assistant. The LLM's primary role was to help refine sentence structure, 
improve clarity, and ensure grammatical correctness and consistency in tone. All scientific contributions, including the research ideation, methodological design, 
experimental analysis, and interpretation of results, were conducted solely by the authors, who take full responsibility for the content of this work.

\end{document}